\documentclass[final,5p,times,twocolumn]{elsarticle}

\usepackage{amssymb}
\usepackage{amsmath}
\usepackage{cleveref}
\usepackage{booktabs}
\usepackage{subcaption}
\usepackage{multirow}
\usepackage{microtype}
\usepackage{float}
\usepackage{longtable}

\newcommand{\datasetsNew}{
\begin{table}[h!]
\centering
\caption{Summary of dataset characteristics. Datasets are categorized as
\emph{short} ($L<150$), \emph{medium} ($150\leq L\leq500$), and
\emph{long} ($L>500$) according to their original time-series length $L$.
Challenging datasets with a large number of output classes are highlighted in bold.}
\label{tab:datasets}

\begin{tiny}
\setlength{\tabcolsep}{2pt}
\renewcommand{\arraystretch}{0.8}

\begin{tabular}{@{}clrrrrl@{}}
\toprule
Scale & Dataset & Train & Test & Class & Length & Type \\
\midrule

\multirow{23}{*}{Short}
& ItalyPowerDemand                & 67   & 1029 & 2  & 24  & Sensor \\
& \textbf{SyntheticControl}       & \textbf{300} & \textbf{300} & \textbf{6} & \textbf{60} & \textbf{Simulated} \\
& SonyAIBORobotSurface2           & 27   & 953  & 2  & 65  & Sensor \\
& SonyAIBORobotSurface1           & 20   & 601  & 2  & 70  & Sensor \\
& MiddlePhalanxOutlineCorrect     & 600  & 291  & 2  & 80  & Image \\
& MiddlePhalanxOutlineAgeGroup    & 400  & 154  & 3  & 80  & Image \\
& DistalPhalanxOutlineCorrect     & 600  & 276  & 2  & 80  & Image \\
& DistalPhalanxOutlineAgeGroup    & 400  & 139  & 3  & 80  & Image \\
& ProximalPhalanxOutlineAgeGroup  & 400  & 205  & 3  & 80  & Image \\
& ProximalPhalanxOutlineCorrect   & 600  & 291  & 2  & 80  & Image \\
& PhalangesOutlinesCorrect        & 1800 & 858  & 2  & 80  & Image \\
& TwoLeadECG                      & 23   & 1139 & 2  & 82  & ECG \\
& MoteStrain                      & 20   & 1252 & 2  & 84  & Sensor \\
& ECG200                          & 100  & 100  & 2  & 96  & ECG \\
& \textbf{ElectricDevices}        & \textbf{8926} & \textbf{7711} & \textbf{7} & \textbf{96} & \textbf{Device} \\
& \textbf{MedicalImages}          & \textbf{381} & \textbf{760} & \textbf{10} & \textbf{99} & \textbf{Image} \\
& CBF                             & 30   & 900  & 3  & 128 & Simulated \\
& TwoPatterns                     & 1000 & 4000 & 4  & 128 & Simulated \\
& \textbf{SwedishLeaf}            & \textbf{500} & \textbf{625} & \textbf{15} & \textbf{128} & \textbf{Image} \\
& \textbf{FaceAll}                & \textbf{560} & \textbf{1690} & \textbf{14} & \textbf{131} & \textbf{Image} \\
& \textbf{FacesUCR}               & \textbf{200} & \textbf{2050} & \textbf{14} & \textbf{131} & \textbf{Image} \\
& ECGFiveDays                     & 23   & 861  & 2  & 136 & ECG \\
& \textbf{Plane}                  & \textbf{105} & \textbf{105} & \textbf{7} & \textbf{144} & \textbf{Sensor} \\
\hline

\multirow{17}{*}{Medium}
& GunPoint                        & 50   & 150  & 2  & 150 & Motion \\
& Wafer                           & 1000 & 6164 & 2  & 152 & Sensor \\
& \textbf{Adiac}                  & \textbf{390} & \textbf{391} & \textbf{37} & \textbf{176} & \textbf{Image} \\
& Strawberry                      & 613  & 370  & 2  & 235 & Spectro \\
& Trace                           & 100  & 100  & 4  & 275 & Sensor \\
& Coffee                          & 28   & 28   & 2  & 286 & Spectro \\
& \textbf{uWaveGestureLibraryX}   & \textbf{896} & \textbf{3582} & \textbf{8} & \textbf{315} & \textbf{Motion} \\
& \textbf{uWaveGestureLibraryY}   & \textbf{896} & \textbf{3582} & \textbf{8} & \textbf{315} & \textbf{Motion} \\
& \textbf{uWaveGestureLibraryZ}   & \textbf{896} & \textbf{3582} & \textbf{8} & \textbf{315} & \textbf{Motion} \\
& \textbf{Lightning7}             & \textbf{70} & \textbf{73} & \textbf{7} & \textbf{319} & \textbf{Sensor} \\
& ToeSegmentation2                & 36   & 130  & 2  & 343 & Sensor \\
& Yoga                            & 300  & 3000 & 2  & 426 & Image \\
& \textbf{OSULeaf}                & \textbf{200} & \textbf{242} & \textbf{6} & \textbf{427} & \textbf{Image} \\
& Ham                             & 109  & 105  & 2  & 431 & Spectro \\
& \textbf{Beef}                   & \textbf{30} & \textbf{30} & \textbf{5} & \textbf{470} & \textbf{Spectro} \\
& FordA                           & 1320 & 3601 & 2  & 500 & Sensor \\
& FordB                           & 810  & 3636 & 2  & 500 & Sensor \\
\hline

\multirow{16}{*}{Long}
& \textbf{ShapesAll}              & \textbf{600} & \textbf{600} & \textbf{60} & \textbf{512} & \textbf{Image} \\
& Herring                         & 64   & 64   & 2  & 512 & Image \\
& BeetleFly                       & 20   & 20   & 2  & 512 & Image \\
& BirdChicken                     & 20   & 20   & 2  & 512 & Image \\
& Earthquakes                     & 322  & 139  & 2  & 512 & Sensor \\
& OliveOil                        & 30   & 30   & 4  & 570 & Spectro \\
& Lightning2                      & 60   & 61   & 2  & 637 & Sensor \\
& Computers                       & 250  & 250  & 2  & 720 & Device \\
& SmallKitchenAppliances          & 375  & 375  & 3  & 720 & Device \\
& LargeKitchenAppliances          & 375  & 375  & 3  & 720 & Device \\
& \textbf{NonInvasiveFetalECGThorax1}
& \textbf{1800} & \textbf{1965} & \textbf{42} & \textbf{750} & \textbf{ECG} \\
& \textbf{NonInvasiveFetalECGThorax2}
& \textbf{1800} & \textbf{1965} & \textbf{42} & \textbf{750} & \textbf{ECG} \\
& WormsTwoClass                   & 181  & 77   & 2  & 900 & Motion \\
& \textbf{UWaveGestureLibraryAll}
& \textbf{896} & \textbf{3582} & \textbf{8} & \textbf{945} & \textbf{Motion} \\
& StarLightCurves                 & 1000 & 8236 & 3  & 1024 & Sensor \\
& HandOutlines                    & 1000 & 370  & 2  & 2709 & Image \\
\bottomrule

\end{tabular}
\end{tiny}
\end{table}
}

\newcommand{\diffarchitect}{
\begin{table}[h!]
\centering
\caption{Performance comparison of the teacher, student-base, and student-MemKD models across various student architecture settings on 28 datasets.}
\label{tab:diff_architect}
\begin{scriptsize}
\setlength{\tabcolsep}{3pt} 
\renewcommand{\arraystretch}{1} 
\begin{tabular}{@{}l|ccc|ccc|ccc@{}} \toprule
Model & \multicolumn{3}{c|}{Win} & \multicolumn{3}{c|}{Avg. rank} & \multicolumn{3}{c}{Avg. AUC-PRC} \\ 
 & Teacher & Base & Ours & Teacher & Base & Ours & Teacher & Base & Ours \\ \midrule
SL1\_4 & \textbf{13} & 3 & 12 & 1.86 & 2.57 & \textbf{1.57} & \textbf{73.19} & 61.52 & 69.34 \\
SL1\_8 & 9 & 0 & \textbf{19} & 1.86 & 2.82 & \textbf{1.32} & 73.19 & 65.5 & \textbf{74.78} \\
SL1\_16 & 7 & 0 & \textbf{21} & 2 & 2.75 & \textbf{1.25} & 73.19 & 67.66 & \textbf{76.19} \\
SL2\_16 & 8 & 1 & \textbf{19} & 2.11 & 2.57 & \textbf{1.32} & 73.19 & 70.62 & \textbf{76.43} \\
SL1\_32 & 10 & 2 & \textbf{16} & 2.07 & 2.5 & \textbf{1.43} & 73.19 & 70.65 & \textbf{76.15} \\
SL2\_32 & 3 & 0 & \textbf{25} & 2.25 & 2.64 & \textbf{1.11} & 73.19 & 70.18 & \textbf{78.59} \\
SL1\_64 & 5 & 1 & \textbf{22} & 2.32 & 2.43 & \textbf{1.25} & 73.19 & 72.42 & \textbf{77.89} \\
SL2\_64 & 2 & 0 & \textbf{26} & 2.36 & 2.57 & \textbf{1.07} & 73.19 & 72.69 & \textbf{79.57} \\
SL3\_64 & 1 & 1 & \textbf{26} & 2.43 & 2.46 & \textbf{1.07} & 73.19 & 72.27 & \textbf{80.48} \\ \bottomrule
\end{tabular}
\end{scriptsize}
\end{table}
}

\newcommand{\networkParams}{
\begin{table}[h!]
\centering
\caption{Configurations of teacher and student networks}
\label{tab:teacher_student_config}
\begin{scriptsize}
\setlength{\tabcolsep}{2pt} 
\renewcommand{\arraystretch}{1} 
\begin{tabular}{@{}lcccccc@{}} \\ \toprule
Model & \begin{tabular}[c]{@{}c@{}}Layer1\\ (hidden size)\end{tabular} & \begin{tabular}[c]{@{}c@{}}Layer2\\ (hidden size)\end{tabular} & \begin{tabular}[c]{@{}c@{}}Layer3\\ (hidden size)\end{tabular} & \begin{tabular}[c]{@{}c@{}}Total \\ param.\end{tabular} & \begin{tabular}[c]{@{}c@{}}Model \\ size(MB)\end{tabular} & \begin{tabular}[c]{@{}c@{}}Compression\\  ratio\end{tabular} \\ \midrule
SL1\_4 & 4 & - & - & 488 & 0.0005 & 0.0006 \\
SL1\_8 & 8 & - & - & 1480 & 0.0014 & 0.0018 \\
SL1\_16 & 16 & - & - & 5000 & 0.0048 & 0.0062 \\
SL2\_16 & 16 & 16 & - & 13704 & 0.0131 & 0.0169 \\
SL1\_32 & 32 & - & - & 18184 & 0.0173 & 0.0224 \\
SL2\_32 & 32 & 32 & - & 51976 & 0.0496 & 0.064 \\
SL1\_64 & 64 & - & - & 69128 & 0.0659 & 0.0851 \\
SL2\_64 & 64 & 64 & - & 202248 & 0.1929 & 0.2491 \\
SL3\_64 & 64 & 64 & 64 & 335368 & 0.3198 & 0.413 \\
Teacher & 100 & 100 & 100 & 812008 & 0.7744 & 1 \\ \bottomrule
\end{tabular}
\end{scriptsize}
\end{table}
}

\newcommand{\ablation}{
\begin{table}[h!]
\centering
\caption{Ablation study on MemKD loss terms}
\label{tab:ablation}
\begin{scriptsize}
\setlength{\tabcolsep}{2pt} 
\renewcommand{\arraystretch}{1} 
\begin{tabular}{@{}cc|ccccccc@{}} \toprule
\multicolumn{2}{c|}{Componenets} & \multicolumn{7}{c}{Metrics} \\ \midrule
\begin{tabular}[c]{@{}c@{}}MemKD-\\ Long\end{tabular} & \begin{tabular}[c]{@{}c@{}}MemKD-\\ Short\end{tabular} & \begin{tabular}[c]{@{}c@{}}Avg. \\ AUC-PRC\end{tabular} & \begin{tabular}[c]{@{}c@{}}Avg. \\ AUC-ROC\end{tabular} & \begin{tabular}[c]{@{}c@{}}Avg. \\ Acc.\end{tabular} & Wins & Tie & Lose & \begin{tabular}[c]{@{}c@{}}Avg. \\ Rank\end{tabular} \\ \midrule
 &  & 65.5 & 78.01 & 62.67 & 0 & 0 & 28 & 3.96 \\
 & x & 72.24 & 83.22 & 67.4 & 6 & 0 & 22 & 2.43 \\
x &  & 74.04 & 84.24 & 68.6 & 7 & 0 & 21 & 2.07 \\
x & x & \textbf{74.78} & \textbf{84.89} & \textbf{70.06} & \textbf{15} & \textbf{0} & \textbf{13} & \textbf{1.54} \\ \bottomrule
\end{tabular}
\end{scriptsize}
\end{table}
}

\newcommand{\BasevsOursCombined}{
\begin{table}[h!]
\centering
\caption{Performance comparison of our method with baseline methods across 56 UCR datasets.}
\begin{subtable}{0.5\textwidth}
\centering
\caption{Comparison with student-base and base-KD}
\begin{scriptsize}
\setlength{\tabcolsep}{2.4pt} 
\begin{tabular}{@{}lccc@{}}
\toprule
 & Base & Base-KD & Ours \\ 
\midrule
Avg. AUC-PRC & 63.89 & 68.65 & \textbf{74.05} \\
Avg. AUC-ROC & 77 & 80.36 & \textbf{84.65} \\
Avg. Acc. & 62.94 & 67.23 & \textbf{71.41} \\
Win & 0 & 7 & \textbf{49} \\
Tie & 0 & 0 & \textbf{0} \\
Lose & 56 & 49 & \textbf{7} \\
Avg. Rank & 2.89 & 1.98 & \textbf{1.13} \\ 
\bottomrule
\end{tabular}
\end{scriptsize}
\label{tab:base_vs_ours}
\end{subtable}
\hfill
\hfill
\hfill\hfill\hfill
\begin{subtable}{0.5\textwidth}
\centering
\caption{Comparison with Fitnets\\ }
\begin{scriptsize}
\setlength{\tabcolsep}{2.4pt} 
\begin{tabular}{@{}lcc@{}}
\toprule
 & Fitnet & Ours \\ 
\midrule
Avg. AUC-PRC & 68.39 & \textbf{74.05} \\
Avg. AUC-ROC & 80.23 & \textbf{84.65} \\
Avg. Acc. & 66.85 & \textbf{71.41} \\
Win & 5 & \textbf{51} \\
Tie & 0 & \textbf{0} \\
Lose & 51 & \textbf{5} \\
Avg. Rank & 1.91 & \textbf{1.09} \\ 
\bottomrule
\end{tabular}
\end{scriptsize}
\label{tab:fits_vs_ours}
\end{subtable}
\hfill
\label{tab:base_vs_ours_combo}
\end{table}
}

\newcommand{\AUCPlots}{
\begin{figure*}
  \centering
  \begin{subfigure}[t]{0.325\linewidth}
    \includegraphics[width=\textwidth]{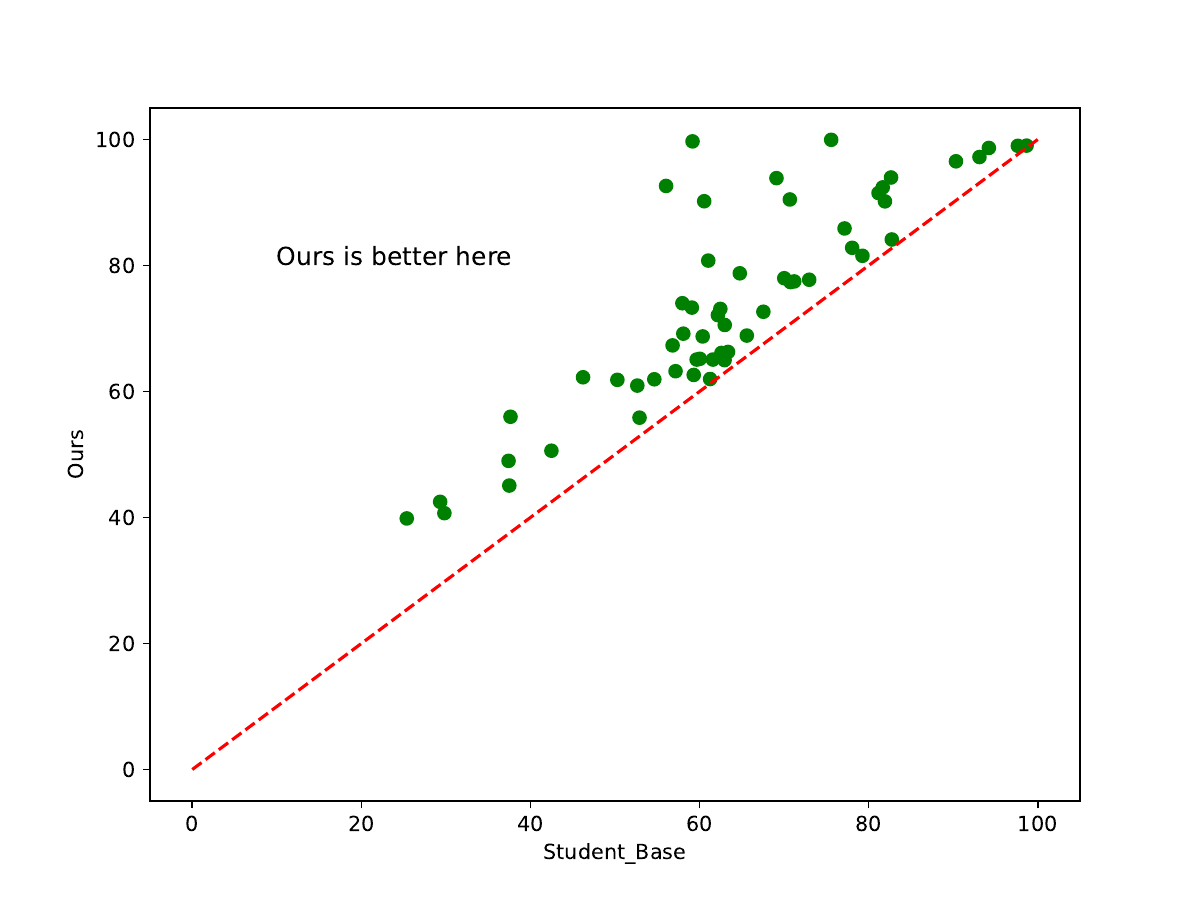}
    \caption{MemKD vs baseline model trained from scratch without knowledge distillation.}
    \label{fig:base_ours}
  \end{subfigure}
  \hfill
  \begin{subfigure}[t]{0.325\linewidth}
    \includegraphics[width=\textwidth]{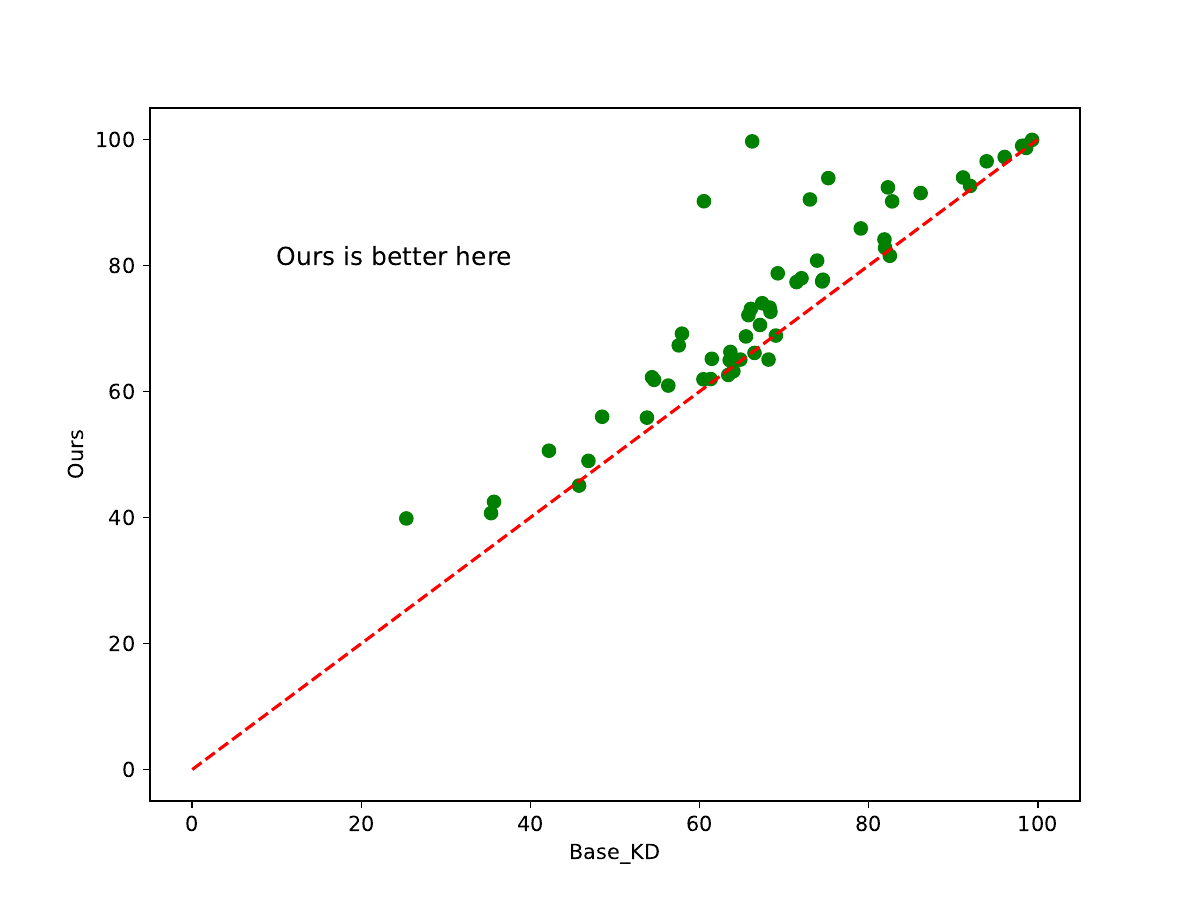}
    \caption{MemKD vs vanilla knowledge distillation (base-KD).}
    \label{fig:hkd_ours}
  \end{subfigure}
  \hfill
  \begin{subfigure}[t]{0.325\linewidth}
    \includegraphics[width=\textwidth]{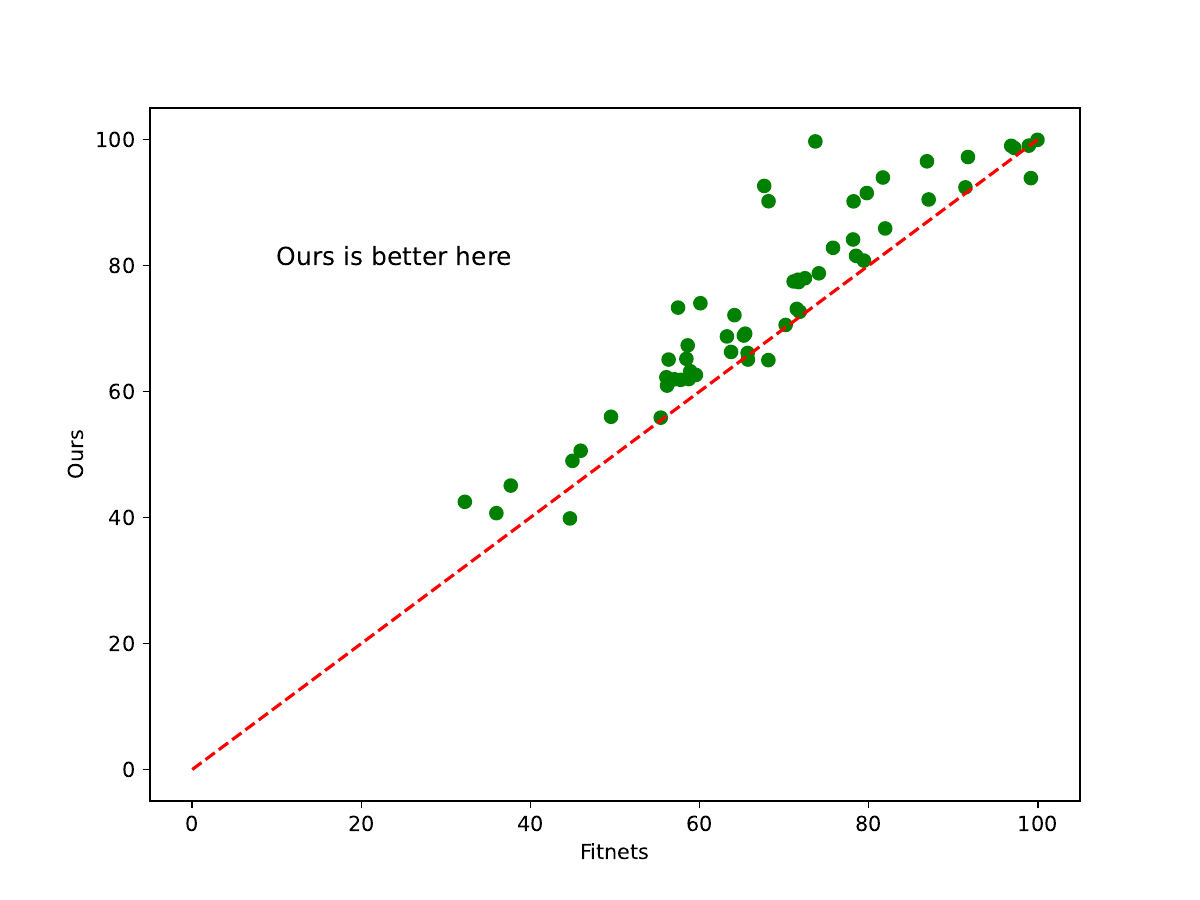}
    \caption{MemKD vs vanilla feature distillation (Fitnets)}
    \label{fig:fits_ours}
  \end{subfigure}
  \caption{AUC-PRC plot showing how the student-MemKD is performing better than baseline model}
  \label{fig:base_all_ours}
  \vspace{-0.15in}
\end{figure*}
}

\newcommand{\rnn}{
\begin{figure}[h!]
\centerline{\includegraphics[width=0.4\textwidth]{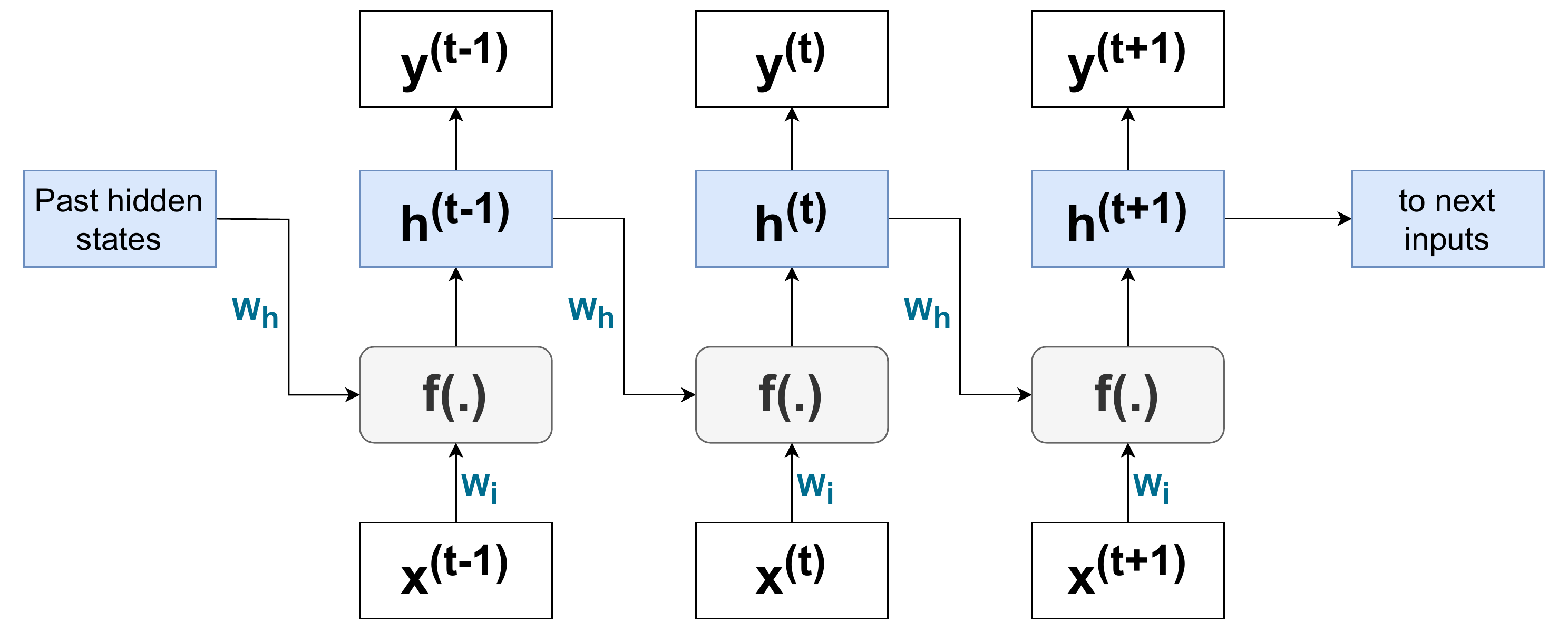}}
  \caption{An overview illustration of a vanilla RNN. The figure depicts the basic structure and flow of information in an RNN that consists of an input layer, hidden layer and an output layer}
  \label{pic:vanilaRNN}
    \vspace{-0.15in}
\end{figure}
}

\newcommand{\overview}{
\begin{figure}[h!]
\centerline{\includegraphics[width=0.4\textwidth]{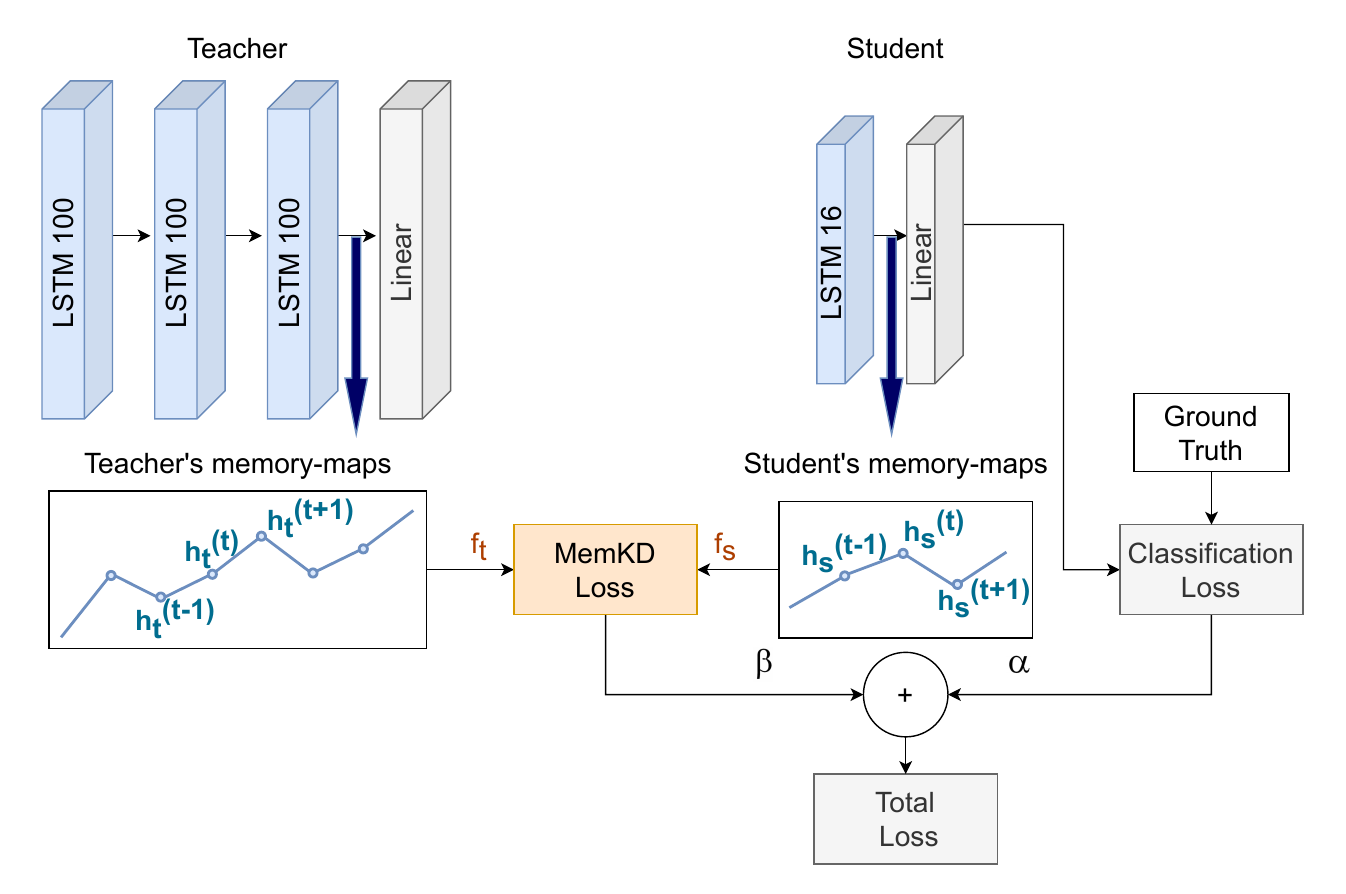}}
  \caption{An overview of the proposed knowledge distillation framework. The student network learns target task by minimizing the classification loss while mimicking the memory retention of the teacher network using intermediate memory-maps.}
  \label{pic:overview}
\end{figure}
}

\newcommand{\sensitivity}{
\begin{figure}[h!]
\centerline{\includegraphics[width=0.4\textwidth]{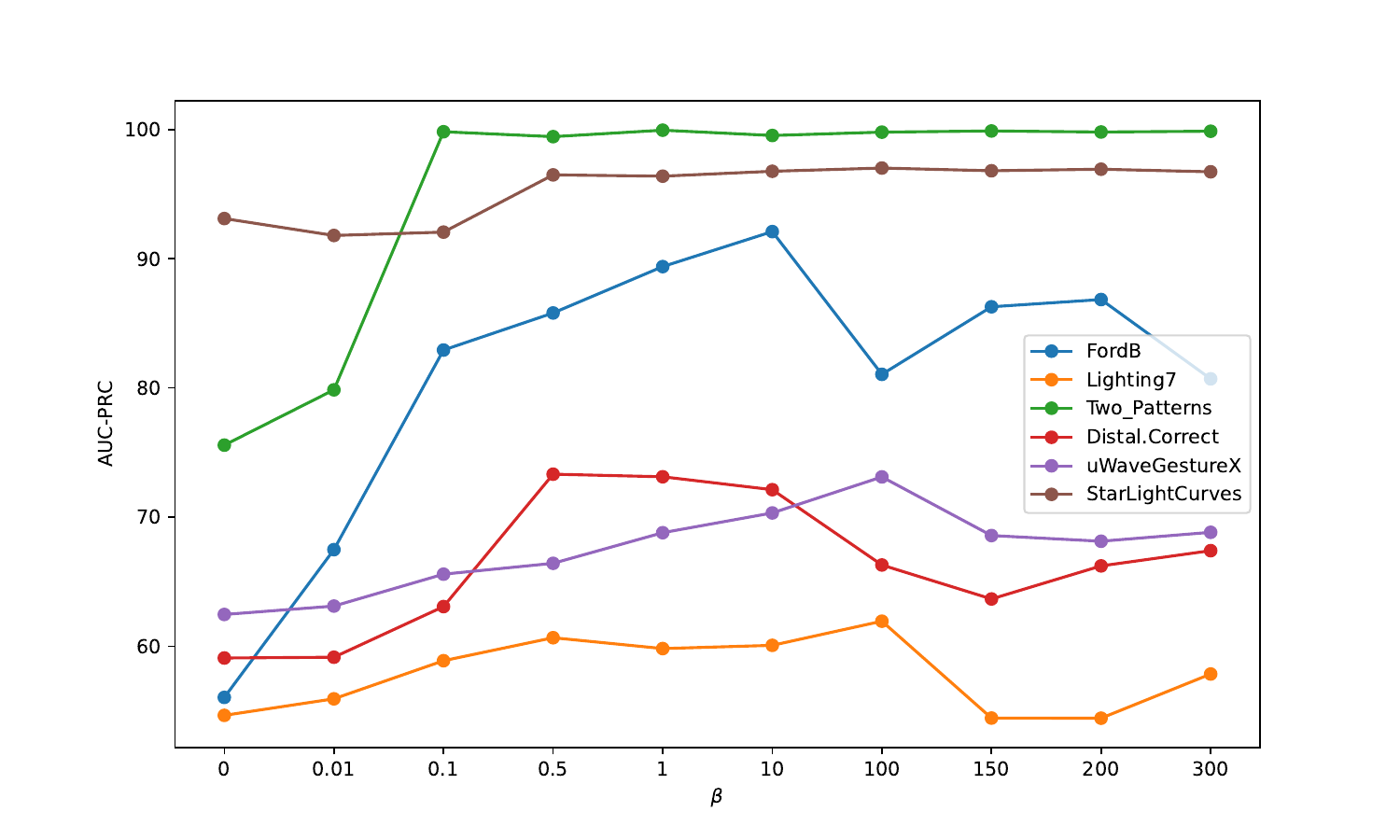}}
  \caption{Sensitivity analysis on $\beta$ for six different datasets. The AUC-PRC values are shown for different $\beta$ values ($\alpha=1$).}
  \label{pic:sensitivity}
    \vspace{-0.15in}
\end{figure}
}

\newcommand{\TSNEPlots}{
\begin{figure}[h!]
  \centering
    \begin{subfigure}{0.90\linewidth}
    \includegraphics[width=\textwidth]{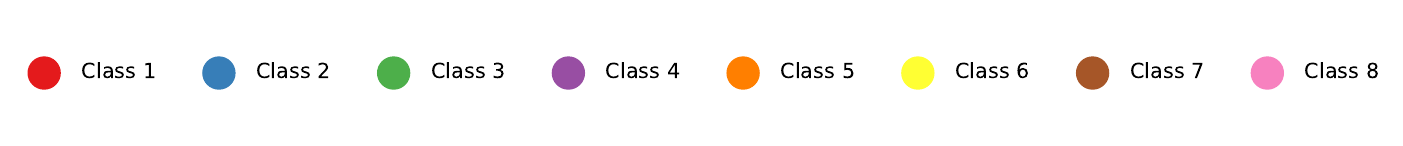}
    \vspace{-0.3in}
    \label{fig:legend}
  \end{subfigure}
  \hfill
  \begin{subfigure}[t]{0.45\linewidth}
    \includegraphics[width=\textwidth]{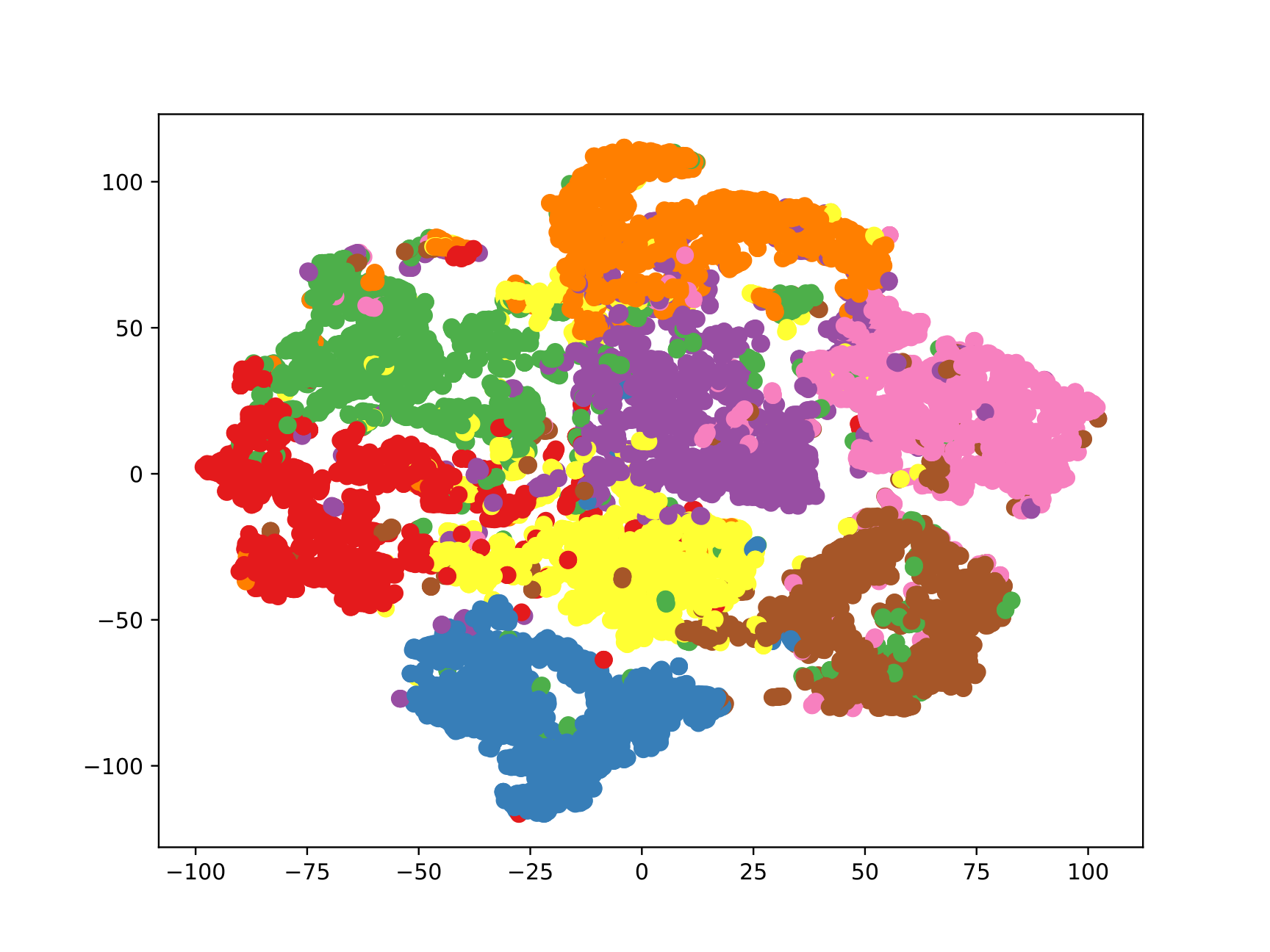}
    \caption{Teacher.}
    \label{fig:teacher-a}
  \end{subfigure}
  \hfill
  \begin{subfigure}[t]{0.45\linewidth}
    \includegraphics[width=\textwidth]{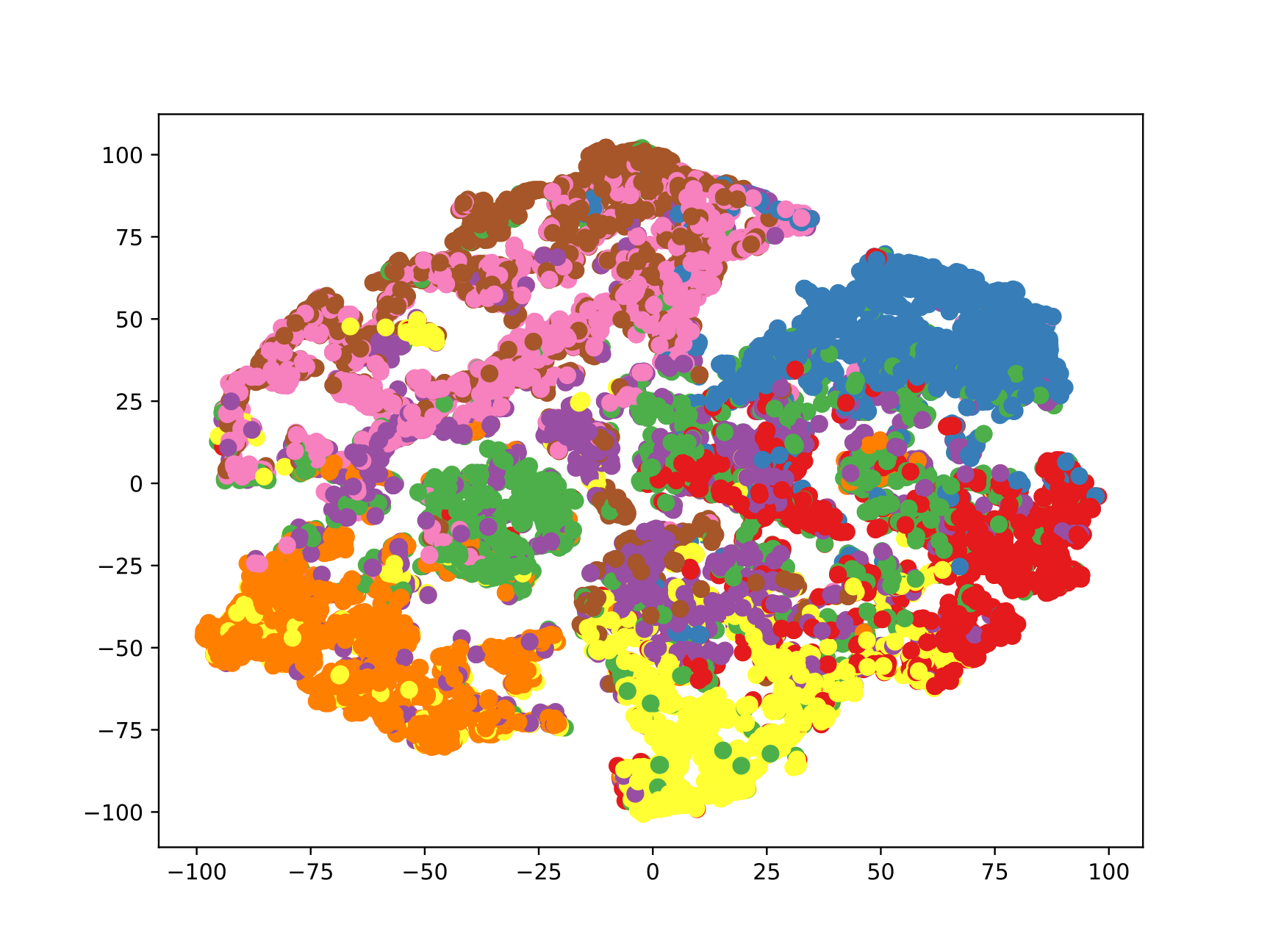}
    \caption{Base.}
    \label{fig:base-b}
  \end{subfigure}
    \hfill
  \begin{subfigure}[t]{0.45\linewidth}
    \includegraphics[width=\textwidth]{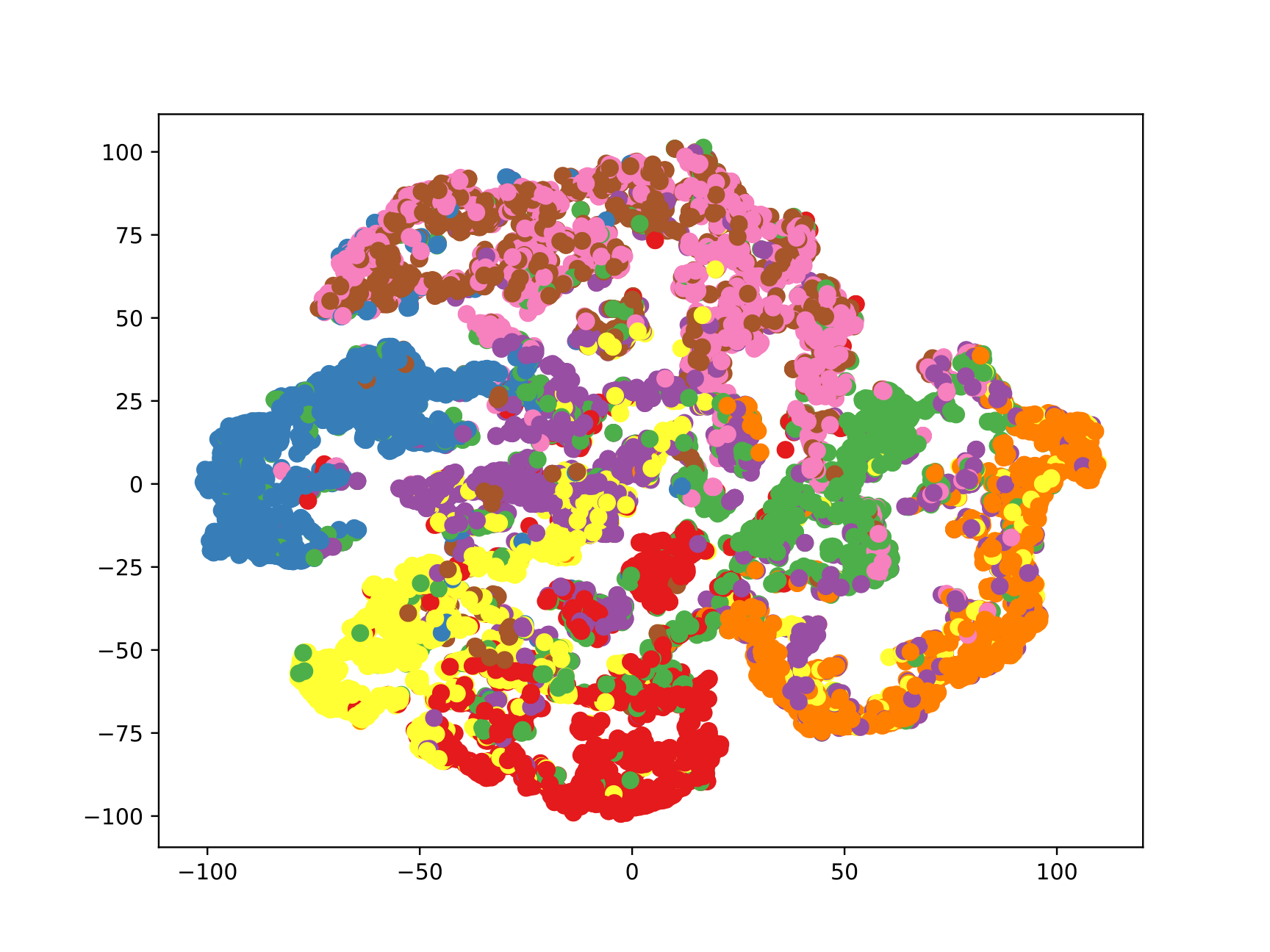}
    \caption{Base-KD.}
    \label{fig:base_kd-c}
  \end{subfigure}
    \hfill
  \begin{subfigure}[t]{0.45\linewidth}
    \includegraphics[width=\textwidth]{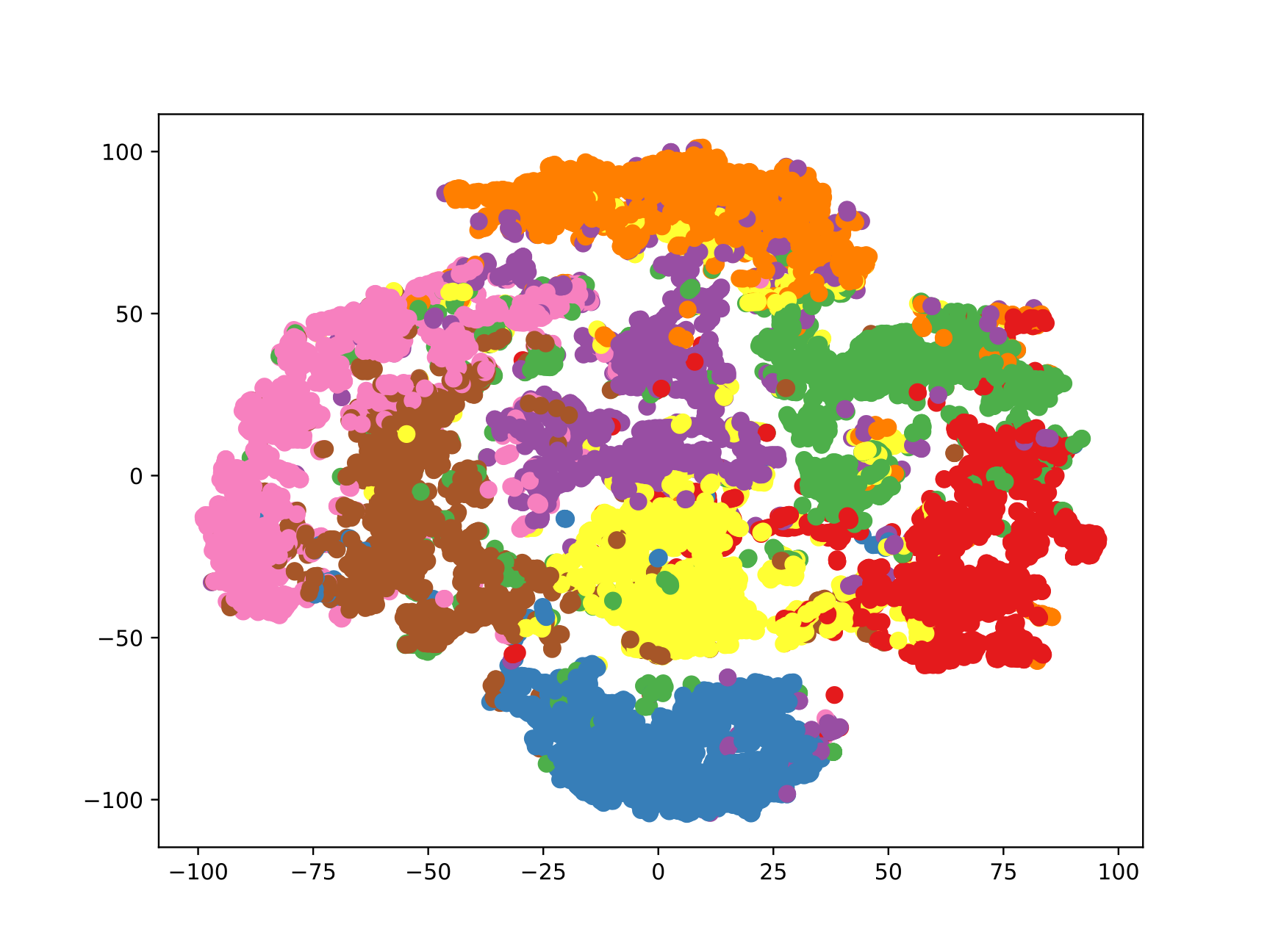}
    \caption{MemKD.}
    \label{fig:ours-d}
  \end{subfigure}
  \caption{Feature space representations of the test set for the UWaveGestureLibraryAll dataset using t-SNE. The last five hidden states from all models were included as candidate features. SL2\_16 models were used as the student models.}
  \label{fig:tsne_plots}
  \vspace{-0.15in}
\end{figure}
}

\newcommand{\numPairs}{
\begin{table}[h!]
\centering
\caption{Sensitivity analysis of the number of chosen subsequences on MemKD-Long}
\begin{scriptsize}
\setlength{\tabcolsep}{2pt} 
\renewcommand{\arraystretch}{1} 
\begin{tabular}{@{}cccc@{}}
\toprule
\begin{tabular}[c]{@{}c@{}}Num. of \\ subseq.\end{tabular} & \begin{tabular}[c]{@{}c@{}}Avg. \\ AUC-PRC\end{tabular} & \begin{tabular}[c]{@{}c@{}}Avg. \\ AUC-ROC\end{tabular} & \begin{tabular}[c]{@{}c@{}}Avg. \\ ACC\end{tabular} \\ \midrule
50 & 73.52 & 83.71 & 69.37 \\
100 & 74.04 & 84.24 & 68.6 \\
150 & 73.83 & 84.18 & 68.87 \\
200 & 74.21 & 84.46 & 69.38 \\
300 & 74 & 84.59 & 69 \\ \bottomrule
\end{tabular}
\end{scriptsize}
\label{tab:num_pairs}
\end{table}
}

\newcommand{\sotaFiftySixNew}{
\begin{table}[h!]
\centering
\caption{Comparison of results with state-of-the-art methods.}
\begin{tiny}
\setlength{\tabcolsep}{2pt} 
\renewcommand{\arraystretch}{0.65} 
\resizebox{\linewidth}{!}{%
\begin{tabular}{@{}lccccccccccc@{}}
\toprule
Dataset & Teacher & Base & \begin{tabular}[c]{@{}c@{}}Base\\ -KD\end{tabular} & DKD & Att. & \begin{tabular}[c]{@{}c@{}}RKD\\ -D\end{tabular} & \begin{tabular}[c]{@{}c@{}}RKD\\ -A\end{tabular} & \begin{tabular}[c]{@{}c@{}}RKD\\ -DA\end{tabular} & Fitnet & DT2W & Ours \\ \midrule
Computers & 59.58 & 57.16 & 64 & 63.78 & 63.72 & 62.66 & \textbf{64.12} & 62.76 & 58.9 & 62.17 & 63.22 \\
UWaveGestureLib.All & 91.72 & 77.15 & 79.07 & 80.58 & 79.31 & 78.34 & 78.99 & 77.51 & 81.96 & 81.35 & \textbf{85.88} \\
Strawberry & 87.91 & 70.68 & 73.06 & 71.13 & 89.09 & 90.05 & 91.53 & \textbf{92.08} & 87.1 & 59.54 & 90.48 \\
BeetleFly & 81.64 & 81.17 & 86.15 & 84.16 & 88.06 & 90.98 & \textbf{93.57} & 90.03 & 79.78 & 76.99 & 91.5 \\
wafer & 81.19 & 98.68 & 98.55 & 98.83 & 98.98 & 98.7 & 98.81 & 98.72 & 98.95 & 89.35 & \textbf{99.03} \\
CBF & 72.44 & 69.1 & 75.23 & 90.07 & 76.6 & 89.74 & 96.29 & 93 & \textbf{99.19} & 51.29 & 93.87 \\
Adiac & 65.79 & 37.5 & 45.75 & 46.36 & \textbf{49.21} & 41.93 & 43.3 & 39.94 & 37.67 & 43.66 & 45.07 \\
Lighting2 & 61.84 & 65.59 & 69.03 & 68.24 & 69.65 & 68.29 & \textbf{70.01} & 68.81 & 65.24 & 66.73 & 68.88 \\
ItalyPowerDemand & 98.21 & 94.22 & 98.63 & 97.68 & 97.55 & 98.37 & \textbf{99.07} & 98.78 & 97.23 & 97.94 & 98.66 \\
yoga & 73.04 & 57.97 & 67.41 & 66.06 & 70.61 & 68.36 & 67.78 & 69.33 & 60.1 & 52.12 & \textbf{74.01} \\
Trace & 54.39 & 64.76 & 69.25 & 68.21 & 61.61 & 72.71 & 72.94 & 72.48 & 74.11 & 53.14 & \textbf{78.76} \\
ShapesAll & 71.04 & 29.82 & 35.34 & \textbf{44.46} & 35.11 & 30.27 & 30.98 & 31.98 & 35.97 & 33.55 & 40.69 \\
Beef & 50.56 & 37.41 & 46.86 & 44.03 & 49.53 & 44.71 & 48.95 & \textbf{50.65} & 44.97 & 37.35 & 48.99 \\
Herring & 55.6 & 50.28 & 54.62 & 57.06 & 61.01 & 56.36 & 55.25 & 55.77 & 57.77 & 58 & \textbf{61.85} \\
MiddlePhalanx.Correct & 62.03 & 60.37 & 65.49 & 67.18 & 65.54 & 66.48 & 68.21 & 66.79 & 63.24 & 63.17 & \textbf{68.74} \\
FordA & 65.66 & 90.33 & 93.96 & 93.2 & 95.58 & 96.34 & 96.29 & 96.12 & 86.9 & 49.73 & \textbf{96.54} \\
SwedishLeaf & 89.65 & 70.75 & 71.47 & 72.95 & 76.39 & 70.87 & 72.94 & 71.11 & 71.72 & 73.14 & \textbf{77.37} \\
FaceAll & 82.77 & 46.22 & 54.38 & 59.82 & 58.21 & 52.68 & 54.6 & 57.88 & 56.08 & 55.92 & \textbf{62.26} \\
StarLightCurves & 96.8 & 93.11 & 96.09 & 95.59 & 95.54 & 95.08 & 93.63 & 94.37 & 91.75 & 92.19 & \textbf{97.22} \\
ECG200 & 69.84 & 67.55 & 68.39 & 68.5 & 70.35 & 73.09 & \textbf{73.79} & 73.07 & 71.85 & 66.1 & 72.66 \\
ECGFiveDays & 83.32 & 81.67 & 82.28 & 83.33 & 91.57 & 85.32 & 89.02 & 85.96 & 91.45 & 76.7 & \textbf{92.4} \\
OliveOil & 36.15 & 25.37 & 25.32 & 25.06 & 32 & 34.38 & 37.76 & 35 & \textbf{44.67} & 37.93 & 39.85 \\
MoteStrain & 80.75 & 82.74 & 81.87 & 81.39 & 82.58 & 83.02 & 83.53 & 82.76 & 78.16 & \textbf{84.21} & 84.13 \\
SonyAIBORobotSurface & 67.2 & 62.96 & 63.57 & 63.23 & 65.55 & 66.74 & \textbf{70.76} & 69.49 & 68.14 & 56.35 & 64.98 \\
SonyAIBORobotSurfaceII & 71.96 & 70.02 & 72.06 & 72.78 & 73.48 & 72.6 & 75.61 & 75.15 & 72.47 & 72.85 & \textbf{77.98} \\
Ham & 69.17 & 62.17 & 65.78 & 65.5 & 70.4 & 68.16 & 69.34 & 68.65 & 64.13 & 57.06 & \textbf{72.12} \\
NonInvasiveFatalECG1 & 82.29 & 58.08 & 57.92 & 56.96 & 64.21 & 61.02 & 58.4 & 55.09 & 65.4 & 55.01 & \textbf{69.18} \\
NonInvasiveFatalECG2 & 86.7 & 71.2 & 74.5 & 66.69 & 73.26 & 73.25 & 73.51 & 73.41 & 71.13 & 72.64 & \textbf{77.47} \\
WormsTwoClass & 51.44 & 52.9 & 53.78 & 53.55 & 55.69 & 56.1 & 56.18 & \textbf{56.53} & 55.41 & 54.16 & 55.85 \\
OSULeaf & 47.4 & 42.48 & 42.19 & 48.59 & 46.51 & 48.46 & 43.67 & 44.72 & 45.94 & 39.62 & \textbf{50.6} \\
uWaveGestureLibrary\_Z & 68.56 & 61.58 & 64.82 & 61.98 & 64.08 & 63.61 & 61.65 & 63.57 & \textbf{65.72} & 65.54 & 65.07 \\
SmallKitchenAppliances & 66.88 & 59.66 & 68.16 & 65.38 & 64.65 & 66.52 & \textbf{70.08} & 65.89 & 56.34 & 59.36 & 65.07 \\
synthetic\_control & 99.51 & 97.65 & 98.16 & 98.3 & 98.83 & 98.7 & 98.69 & 98.69 & 96.86 & \textbf{99.22} & 99 \\
TwoLeadECG & 68.24 & 62.98 & 67.15 & 67.28 & 66.79 & 70.31 & \textbf{72.62} & 68.52 & 70.18 & 52.9 & 70.56 \\
Two\_Patterns & 100 & 75.57 & 99.34 & 99.16 & 99.97 & 99.96 & 99.93 & \textbf{99.99} & 99.97 & 76.04 & 99.95 \\
MiddlePhalanx.AgeGroup & 62.98 & 63.37 & 63.65 & 63.98 & 63.7 & 64.63 & 65.38 & 64.26 & 63.73 & 62.25 & \textbf{66.28} \\
DistalPhalanx.Correct & 69.12 & 59.09 & 68.29 & 69.11 & 63.96 & 70.98 & 68.41 & 66.68 & 57.46 & 59.42 & \textbf{73.32} \\
FacesUCR & 70.73 & 37.64 & 48.48 & 55.38 & 46.72 & 46.96 & 46.09 & 49.36 & 49.53 & 32.24 & \textbf{55.99} \\
HandOutlines & 62.69 & 60.04 & 61.45 & 63.31 & 60.94 & \textbf{72.84} & 71.64 & 69.43 & 58.44 & 60.3 & 65.19 \\
Coffee & 81.18 & 59.17 & 66.23 & 66.23 & 76.15 & 73.71 & 81.42 & 77.99 & 73.7 & 63.37 & \textbf{99.7} \\
Gun\_Point & 75.91 & 60.55 & 60.52 & 61 & 69.15 & 73.7 & 79.31 & 77.35 & 68.16 & 59.52 & \textbf{90.2} \\
FordB & 75.45 & 56.04 & 92.01 & 91.75 & 91.27 & 91 & 91.88 & \textbf{93.03} & 67.65 & 51.19 & 92.63 \\
MedicalImages & 40.87 & 29.32 & 35.69 & 31.91 & 41.09 & 29.84 & 33.73 & 31.33 & 32.24 & 27.49 & \textbf{42.49} \\
ProximalPhalanx.AgeGroup & 80.92 & 79.27 & 82.51 & 80.94 & 82.6 & \textbf{83.05} & 81.87 & 81.86 & 78.51 & 78.76 & 81.54 \\
Plane & 73.75 & 82.65 & 91.17 & 97.44 & 97.07 & 96.38 & \textbf{97.91} & 95.89 & 81.69 & 84.16 & 93.97 \\
Lighting7 & 57.43 & 54.65 & 60.46 & 57.98 & 60.21 & \textbf{62.82} & 59.82 & 58.87 & 57.02 & 52.38 & 61.94 \\
Phalanges.Correct & 75.51 & 72.97 & 74.59 & 75.07 & 76.99 & 76.19 & 75.83 & 75.19 & 71.64 & 69.73 & \textbf{77.74} \\
DistalPhalanx.AgeGroup & 82.39 & 78.05 & 81.95 & 81.09 & 82.11 & 80.09 & 80.1 & 82.04 & 75.79 & 69.43 & \textbf{82.81} \\
uWaveGestureLibrary\_X & 76.68 & 62.46 & 66.07 & 71.96 & 67.79 & 65.08 & 64.95 & 64.54 & 71.5 & 69.85 & \textbf{73.11} \\
ToeSegmentation2 & 72.04 & 52.64 & 56.3 & 58.14 & 56.88 & 64.54 & \textbf{69.51} & 62.96 & 56.16 & 57.87 & 60.94 \\
BirdChicken & 52.51 & 56.81 & 57.54 & 57.54 & 61.13 & 66.75 & 60.8 & 65.68 & 58.6 & 55.19 & \textbf{67.32} \\
LargeKitchenApp & 75.51 & 61.03 & 73.91 & 80.45 & 77.43 & 82.14 & \textbf{84.21} & 81.09 & 79.44 & 65.98 & 80.78 \\
uWaveGestureLibrary\_Y & 66.94 & 59.31 & 63.4 & \textbf{64.42} & 62.37 & 59.65 & 60.35 & 60.79 & 59.57 & 61.48 & 62.63 \\
ProximalPhalanx.Correct & 85.51 & 81.93 & 82.78 & 82.5 & 89.85 & 88.25 & 85.58 & 86.72 & 78.22 & 78.6 & \textbf{90.18} \\
ElectricDevices & 66.48 & 62.59 & \textbf{66.5} & 63.91 & 65.64 & 64.63 & 65.27 & 65.78 & 65.69 & 65.9 & 66.12 \\
Earthquakes & 55.25 & 61.25 & 61.32 & 61.92 & 61.82 & 62.05 & \textbf{62.11} & 61.41 & 58.73 & 61.86 & 61.98 \\ \hline
 & & \textbf{} & \textbf{} & \textbf{} & \textbf{} & \textbf{} & \textbf{} & \textbf{} & \textbf{} & \textbf{} & \textbf{} \\ 
Avg. AUC-PRC & 71.63 & 63.89 & 68.65 & 69.34 & 70.64 & 70.88 & 71.75 & 71.02 & 68.39 & 62.68 & \textbf{74.05} \\
Avg. AUC-ROC & & 77.00 & 80.36 & 80.83 & 82.04 & 82.82 & 83.65 & 82.78 & 80.23 & 75.30 & \textbf{84.65} \\
Avg. Acc. & & 62.94 & 67.23 & 67.55 & 69.25 & 68.80 & 70.19 & 69.25 & 66.85 & 62.12 & \textbf{71.41} \\
Wins & \_\_ & 0 & 1 & 2 & 1 & 3 & 12 & 5 & 3 & 2 & \textbf{27} \\
Lose & \_\_ & 56 & 55 & 54 & 55 & 53 & 44 & 51 & 53 & 54 & \textbf{29} \\
Avg. rank & \_\_ & 8.98 & 5.95 & 5.68 & 4.52 & 4.8 & 3.89 & 4.59 & 6.5 & 7.73 & \textbf{2.29} \\
Wins(with teacher) & 14 & 0 & 1 & 0 & 0 & 3 & 11 & 4 & 2 & 1 & \textbf{20} \\
Lose(with teacher) & 42 & 56 & 55 & 56 & 56 & 53 & 45 & 52 & 54 & 55 & \textbf{36} \\
Avg. rank(with teacher) & 5.34 & 9.8 & 6.57 & 6.27 & 5.04 & 5.27 & 4.3 & 5.05 & 7.16 & 8.52 & \textbf{2.61} \\ \bottomrule
\end{tabular}}
\end{tiny}
\label{tab:sota}
\end{table}
}

\journal{Pattern Recognition}

\begin{document}

\begin{frontmatter}



\title{\fontsize{14pt}{17pt}\selectfont
Learning to Remember: Distilling Memory Retention for Compact Recurrent Neural Networks}



\author[label1]{Nilushika Udayangani}

\author[label1]{Kishor Nandakishor}
\author[label1]{Marimuthu Palaniswami}

\affiliation[label1]{
    organization={Department of Electrical and Electronic Engineering, University of Melbourne},
    city={Parkville},
    postcode={3052},
    state={VIC},
    country={Australia}
}


\begin{abstract}
Deep learning models, particularly recurrent neural networks and their variants, such as long short-term memory, have significantly advanced time series analysis. These models capture complex, sequential patterns in time series, enabling real-time assessments. However, their high computational complexity and large model sizes pose challenges for deployment in resource-constrained environments, such as wearable devices and edge computing platforms. Knowledge Distillation (KD) offers a solution by transferring knowledge from a large, complex model (teacher) to a smaller, more efficient model (student), thereby retaining high performance while reducing computational demands. Current KD methods, originally designed for computer vision tasks, neglect the unique temporal dependencies and memory retention characteristics of time series models. To bridge this gap, we propose a novel KD framework termed \textbf{Mem}ory-Discrepancy \textbf{K}nowledge \textbf{D}istillation~(\textbf{MemKD}). MemKD leverages a specialized loss function to capture memory retention discrepancies between the teacher and student models across subsequences within time series data, ensuring that the student model effectively mimics the teacher's behaviour. This approach facilitates the development of compact, high-performing recurrent neural networks suitable for real-time, time series analysis tasks. We provide additional experiments, in-depth theoretical analysis, and insights into the proposed framework across extended time series benchmarks. Our experiments demonstrate that MemKD significantly outperforms state-of-the-art KD methods. Additionally, we demonstrate that it can match the teacher model's performance across a wide range of compression levels, achieving notable reductions in parameter count and memory usage without a significant loss in accuracy.
\end{abstract}




\end{frontmatter}



\section{Introduction}
Time series analysis has become fundamental to modern decision-making systems, enabling critical applications from patient health monitoring
to industrial fault detection and environmental sensing. These applications require a careful balance between achieving high performance in handling complex time series data and delivering real-time predictions, while ensuring the solutions remain lightweight and suitable for deployment on edge devices such as smartphones and wearable sensors. Recurrent neural networks~(RNNs) and their variants, particularly long Short-Term memory networks~(LSTMs) and gated recurrent units~(GRUs), are excel at time series analysis due to their adeptness in sequence modeling.
RNNs maintain memory-like arrays known as hidden states, enabling them to retain and recall long-term dynamics within sequential data. However, as input sequences are processed, hundreds of thousands of hidden states need to be updated with nonlinear functions specified by millions of parameters. Consequently, RNNs share common drawbacks inherent to deep neural networks~(DNNs), such as high computational complexity, large parameter spaces, and high inference latency. These limitations pose challenges for real-world time series applications, which demand low power and space requirements.

Knowledge distillation (KD) transfers knowledge from a large teacher model to a smaller student model, maintaining performance while improving efficiency. Introduced by \citet{ba_deep_2014} and extended by~\citet{hinton_distilling_2014,romero_fitnets_2014} for image classification, KD showed that smaller models could match or outperform deeper ones by matching the teacher's outputs or intermediate features. Despite its broad applicability, KD remains understudied in time series analysis. Ay et al.~\cite{ay_study_2022} conducted the first study on KD for time series classification using fully convolutional networks, following \citet{hinton_distilling_2014} to align the teacher and student output probability distributions. Subsequently, Ouyang et al.~\cite{ouyang_knowledge_2023} introduced a KD framework for TSC, extracting knowledge in both time and frequency domains by aligning output distributions in separate domains. 
\citet{ma_knowledge_2022} specifically explored KD using an RNN teacher, yet similarly relied on directly matching the teacher and student outputs.

However, these studies primarily adapt KD frameworks originally developed for computer vision tasks, without tailoring their loss functions to exploit the unique characteristics of time series data. Time series values, as opposed to visual data, are unidirectionally correlated, and during classification, short contiguous subsequences often contain much of the discriminative information~\cite{parvatharaju_learning_2021,ye_time_2009}. Therefore, adaptions of image approaches for time series may disrupt knowledge of those discriminative subsequences and their correlations.

\overview

Considering that the ability to maintain memory is crucial to the success of RNNs, this raises a natural question: why not leverage knowledge of how the teacher learns to retain memory? This is a problem that none of the aforementioned studies have investigated. This brings us to the main topic of this paper: can a teacher network improve the performance of a student network by providing it with information about what it remembers? To this end, we propose a novel KD framework, termed \textit{Memory-Discrepancy Knowledge Distillation~(MemKD)}, which effectively transfers knowledge from RNNs based on their memory maintenance over input sequences, as shown in ~\Cref{pic:overview}. Within this framework, we design a novel loss function to force the student network to mimic expected changes in the teacher's memory retention over short contiguous subsequences. We empirically validate its effectiveness for TSC tasks across various application domains, demonstrating that MemKD successfully distills knowledge from a large LSTM-RNN model to a compact LSTM-RNN model, notably reducing parameter count and memory usage while maintaining performance comparable to the teacher.
The main contributions of our work include:
\begin{itemize}
\item a novel KD framework to distill knowledge from a larger LSTM-RNN model to a much smaller LSTM-RNN model, enhancing efficiency in TSC across various application domains.
\item a novel loss function designed to capture knowledge on how RNNs restore and retain memory for different shapes of subsequences within a time series sequence.
\item an indirect feature-level knowledge transfer method between RNN models, seamlessly transferring knowledge across varying dimensional intermediate representations without needing to match sizes.
\item extensive experiments to demonstrate that our proposed MemKD significantly outperforms the baseline KD methods.
\end{itemize}
The rest of the paper is organized as follows:~\Cref{section_related_works} reviews related works on KD and time series analysis. \Cref{method} presents the details of our proposed MemKD framework.~\Cref{experiments} introduces our experimental setup, followed by experimental results and comparisons with baseline methods.~\Cref{conculsion} provides an in-depth discussion of the results and their implications for future research.

\section{Background and Related Works}\label{section_related_works}
\subsection{Time Series Analysis}
Many real-world phenomena evolve over time, resulting in data that are naturally collected as sequential observations. 
Consequently, time series data typically manifest various systematic patterns, including trends, seasonalities, cyclic variations and irregular movements~\cite{bock_machine_2021}. However, analyzing time series data poses challenges as it provides only two primary features: the point in time and the observed value itself. Analyzing them as sets of subsequences has been the most intuitive approach for extracting discriminative and interpretable features for further processing. Therefore, most traditional machine learning approaches, including shapelets~\cite{ye_time_2009}, motif discovery~\cite{patel_mining_2002}, discord discovery~\cite{chandola_detecting_2009}, and time series chains~\cite{zhu_time_2018}, relied on subsequence analysis for their success. The enduring success of these methods as stable approaches over decades highlights the importance of preserving subsequence-level knowledge, which we account for in designing our knowledge transfer function.

\subsection{Recurrent Neural Networks}
Time series analysis has drastically advanced with the rapid development of deep learning. Among these developments, RNNs are a widely adopted type of neural network that allow for the modelling of the sequential nature of time series data by storing past context~\cite{elman_1990_finding,husken_recurrent_2003}. LSTM~\cite{hochreiter_1997_long} and GRU~\cite{cho_2014_learning} were introduced as extensions to vanilla RNNs, addressing common drawbacks such as the vanishing gradient problem and the difficulty in learning long-term dependencies. These variants have been extensively studied for time series analysis, across various application domains~\cite{morid_2023_time,lindemann_survey_2021}. However, these models exhibit higher computational complexities and significantly slower processing times due to their cascaded connections~\cite{lei_2017_simple,xu_contrastive_2022}. This limits their applicability in edge deployment and necessitates the need for model compression techniques.

\subsection{Semantics from Hidden States}
The core mechanism of RNNs and their variants is a self-connected hidden state layer that captures complex relationships between observations within a sequence. 
Several studies in NLP have attempted to interpret how RNNs work internally with their memory and whether their hidden states encode meaningful semantic information~\cite{strobelt_visual_2016,strobelt_lstmvis_2018}; however, such interpretations of hidden states remain largely unexplored in the context of time series analysis.
For example, Karpathy et al.~\cite{karpathy_visualizing_2015} found that when RNNs are applied to a sequence of words, some hidden states can be interpreted as language context, such as quotes or punctuation, while others cannot. Ming et al.~\cite{ming_understanding_2017} explored how information stored in hidden states correlates with certain word clouds. 
Despite these findings, limited work has exploited such hidden-state semantics for downstream learning, and, to the best of our knowledge, no prior method has explicitly utilized them as a source of knowledge in a KD framework.

\subsection{Knowledge Distillation}
The essence of KD lies in two primary objectives: firstly, extracting rich knowledge from the teacher network, and secondly, transferring this knowledge to guide the training of a student network.
To ensure the quality of knowledge, various methodologies have been proposed. The most commonly utilized methods are logit-level KD techniques, which extract knowledge from softened output distributions~\cite{hinton_distilling_2014}. DKD~\cite{zhao_decoupled_2022} improves performance by decoupling logit-level knowledge from target and non-target class distributions. Beyond the logits from the last layer, Fitnets~\cite{romero_fitnets_2014} proposed directly matching features from intermediate layers. Transferring feature-level knowledge using attention maps has also been studied~\cite{zagoruyko_paying_2017,huang_like_2017}. VID~\cite{ahn_variational_2019} aims to maximize mutual information between the student and teacher networks by variational information maximization, while RKD~\cite{park_relational_2019} proposed extracting relationships among input data samples. These methods are considered state-of-the-art in KD; however, they were originally developed and tested in the context of computer vision tasks. Consequently, the loss functions employed in most of the aforementioned methods were specifically designed to capture visual features and may not have direct analogies for time series data. Recent works have only begun addressing this gap by adapting these image approaches for time series. For example, the work in ~\cite{ay_study_2022,ouyang_knowledge_2023} studied logit-level KD for time series, and DT2W~\cite{qiao_class-incremental_2023} extends Fitnets~\cite{romero_fitnets_2014} by replacing Euclidean distance with soft-dynamic time warping distance. \citet{xu_kdnet-rul_2022} utilized adversarial learning to compress RNN models into CNNs in their studies focusing on disparate architecture KD methods~\cite{xu_kdnet-rul_2022,xu_contrastive_2022}. 

However, possibility of knowledge extraction from RNN memory states has received limited research attention. A closely related study by \citet{murata_recurrent_2023} considers hidden states for time series forecasting by directly aligning the final hidden states of the teacher and student. However, this approach essentially treats hidden states as intermediate features for direct feature matching, without interpreting or exploiting how the information encoded in these states evolves as memory over the input sequence.
In contrast, we propose a new method for compressing RNN models that extracts meaningful knowledge from how memory states evolve, rather than simply matching the states directly. This approach enables seamless knowledge transfer from a larger RNN-based model to a significantly smaller one without requiring the teacher and student hidden states to have matching dimensions.

\section{Methodology}\label{method}

\subsection{Knowledge Distillation for Time Series Classification}
A time series can be represented as \begin{math}
\boldsymbol{X} = \left\{\boldsymbol{x}^{(1)}, \boldsymbol{x}^{(2)},\cdots, \boldsymbol{x}^{(T)}\right\}
\end{math}, where $\boldsymbol{x}^{(i)} \in \mathbb{R}^n$, $i=1,\cdots,T$ and $\boldsymbol{X}\in \mathbb{R}^{T\times n}$.
A time series dataset $\mathcal{D}$, which includes M samples, is a collection of pairs $(\boldsymbol{X}_i, \boldsymbol{Y}_i), i= 1, \cdots,M$, where $\boldsymbol{Y}_i \in \mathbb{R}^C$ is a one-hot vector representing the class label $c\in [1,C]$ corresponds to $\boldsymbol{X}_i$. Therefore, the TSC problem involves training a classifier $\mathcal{F}$, mapping input $\boldsymbol{X}_i$ to their corresponding label $\boldsymbol{Y}_i$. The goal of the KD is to train a student model (a shallow model) as $\mathcal{F}$ on our time series dataset $\mathcal{D}$ by leveraging the knowledge acquired by a pre-trained teacher model (a deep model). This is achieved by minimizing a distillation loss function that can be typically represented as:
\begin{equation}\label{eq_lkd}
L_{\mathrm{KD}} =\sum _{\boldsymbol{X}_i} \ell\left(f_t(\boldsymbol{X}_i),f_s(\boldsymbol{X}_i)\right)
\end{equation} where $f_t$ and $f_s$ represent functions for teacher and student models respectively and $\ell$ represents a similarity measure used to match the metrics of the teacher and student models.

\subsection{Memory-Discrepancy Knowledge Distillation}
\rnn
\subsubsection{Formulating Knowledge from Hidden States.}
Given an input time series sequence
\begin{math}
\boldsymbol{X}_i = \left\{\boldsymbol{x}^{(1)}, \boldsymbol{x}^{(2)}, \cdots, \boldsymbol{x}^{(T)}\right\}
\end{math}, the vanilla RNN depicted in ~\Cref{pic:vanilaRNN} processes each input time step $\boldsymbol{x}^{(t)} \in \mathbb{R}^n $ to maintain a time-varying memory state known as the hidden state $\boldsymbol{h}^{(t)} \in \mathbb{R}^m$. At time step $t$, the RNN updates hidden state $\boldsymbol{h}^{(t-1)}$ to $\boldsymbol{h}^{(t)}$ using both $\boldsymbol{x}^{(t)}$ and $\boldsymbol{h}^{(t-1)}$ as:

\begin{equation}\label{eq1}
\boldsymbol{h}^{(t)} = f(\boldsymbol{W}_h \boldsymbol{h}^{(t-1)} + \boldsymbol{W}_i \boldsymbol{x}^{(t)})
\end{equation}
where $\boldsymbol{W}_h$ and $\boldsymbol{W}_i$ are weight matrices and $f$ is a non-linear activation function which constrains the range of $\boldsymbol{h}^{(t)}$.
After processing the entire sequence at time step $T$, a linear layer can be trained upon last hidden state $\boldsymbol{h}^{(T)}$ to derive the output probability distribution for classification tasks. 

By unrolling past hidden states in~\Cref{eq1}, each hidden state $\boldsymbol{h}^{(t)}$ at time step t can be expressed as a function of past input values up to time step t, \begin{math}
\left\{\boldsymbol{x}^{(1)}, \boldsymbol{x}^{(2)},\cdots, \boldsymbol{x}^{(t)}\right\}
\end{math} and the initial hidden state value $\boldsymbol{h}^{(0)}$.
In other words, each hidden state vector represents the state of the RNN's memory at a particular time step, incorporating information about the input sequence up to that point. It represents a compact representation of the input sequence's temporal dynamics up to that time step. Therefore, the hidden state can be seen as evolving over time as the RNN processes the input sequence, capturing the temporal dependencies and patterns present in the data. Thus patterns induced from memory hold the potential to serve as a rich source of knowledge in a KD framework.

Let us consider the hidden state $\boldsymbol{h}^{(t+z)}$ at time step $t+z$, $z \in \mathbb{Z}$ . By unrolling each hidden state in~\Cref{eq1} up to $\boldsymbol{h}^{(t)}$, it can be written as
\begin{math}
\boldsymbol{h}^{(t+z)} = f(\boldsymbol{h}^{(t)}, \boldsymbol{x}^{(t+1)}, \boldsymbol{x}^{(t+2)},\cdots, \boldsymbol{x}^{(t+z)})
\end{math} where $f$ is a nonlinear function.
Now consider the difference between any two hidden states $\boldsymbol{h}^{(t+z)}$ and $\boldsymbol{h}^{(t)}$ : 
\begin{equation}\label{eq_delatah}
\Delta \boldsymbol{h}^{t,z} = (\boldsymbol{h}^{(t+z)}-\boldsymbol{h}^{(t)})
\end{equation}

This can be regarded as response of memory to input values in subsequence, \\
\begin{math}
\left\{\boldsymbol{x}^{(t+1)}, \boldsymbol{x}^{(t+2)},\cdots,\boldsymbol{x}^{(t+z-1)}, \boldsymbol{x}^{(t+z)}\right\}
\end{math}, denoted $\boldsymbol{X}^{t:(t+z)}$ here. Although $\boldsymbol{h}^{(t+z)}$ is calculated by a non-linear transformation of $\boldsymbol{h}^{(t)}$ and values in $\boldsymbol{X}^{t:(t+z)}$, $\Delta \boldsymbol{h}^{t,z}$ can be regarded as deterministic to the input values in subsequence $\boldsymbol{X}^{t:(t+z)}$ when the previous history $\boldsymbol{h}^{(t)}$ is given. Thus $\Delta \boldsymbol{h}^{t,z}$ can reflect to what degree the model's hidden state (memory) is influenced by the input values in subsequence $\boldsymbol{X}^{t:(t+z)}$. 

\subsubsection{Knowledge Distillation Objective. }
We hypothesize that, after the occurrence of certain patterns in time series data, the model's tendency to memorize subsequent smaller occurrences or data patterns provides a valuable source of knowledge to extract. Thus we define a novel distillation loss function based on the metric defined in~\Cref{eq_delatah} \textit{i.e.}, the measure of the degree to which an input subsequence influences the model's memory.
The suggested loss function tries to minimize the difference between these metrics
among teacher and student. Explicitly we define $f_t$ and $f_s$ in~\Cref{eq_lkd} as follow:
\begin{equation}\label{eq_ft}
f_t = \left\{\frac{\Delta \boldsymbol{h}_t^{t,z}}{ \parallel \boldsymbol{h}_t^{(t)} \parallel}\right\}t,z\in [T-1], t+z\leq T
\end{equation} 
\begin{equation}\label{eq_fs}
f_s = \left\{\frac{\Delta \boldsymbol{h}_s^{t,z}}{ \parallel \boldsymbol{h}_s^{(t)} \parallel}\right\}t,z\in [T-1],t+z\leq T
\end{equation} 
where $\boldsymbol{h}_t$ and $\boldsymbol{h}_s$ represent hidden state vectors of teacher and student respectively. To mitigate any significant scale differences between the teacher's and student's hidden states and thereby ensure the stability of the loss function, we normalize each difference by $\parallel \boldsymbol{h}^{(t)}\parallel$ in~\Cref{eq_ft} and~\Cref{eq_fs}.
 As for the metric $\ell$ to measure the difference between~\Cref{eq_ft} and~\Cref{eq_fs}, we utilize \textit{Smooth L1 loss}:
\begin{equation}\label{eq_memkd}
L_{\mathrm{MemKD}} = \sum_{\boldsymbol{X}_i} \ell_{\mathrm{smoothL1}}\left(f_t(\boldsymbol{X}_i), f_s(\boldsymbol{X}_i)\right)
\end{equation}
where $L_{\mathrm{MemKD}}$ represents the proposed memory-discrepancy distillation loss.
Smooth L1 loss incorporates a quadratic loss term for small differences and a linear loss term otherwise, making it smoother than L1 loss. It is less sensitive to outliers than L2 loss and helps prevent exploding gradients~\cite{girshick_2015_fast}.

In our study, we analyze the impact of the memory-discrepancy distillation loss formed by shorter and longer subsequences separately to understand their individual contributions more effectively. In the first stage, we fix $z=1$ and consider all $\Delta \boldsymbol{h}^{t,z}$ formed by every pair of consecutive hidden states, i.e. $\left\{\Delta \boldsymbol{h}^{t,1}\right\}_{t=1,\cdots,(T-1)}$. Thus, each $\Delta \boldsymbol{h}^{t,1}$ now represents how much the memory state is affected by current input value $\boldsymbol{x}^{(t)}$ at time step $t \in \{1,\cdots,T\}$, given the $\boldsymbol{X}^{0:(t-1)}$ has occurred. We refer to this as the memory-discrepancy distillation loss for shorter subsequences ($z=1$), denoted as $L_{\mathrm{MemKD-Short}}$. 
In the subsequent stage, we analyze the effects of longer subsequences by selecting larger values of $z$
(where $z>1$). Since it is impractical to form all possible subsequences corresponding to every $t\in[T-1]$ and $z\in [T-1] \setminus \{1\}$, we select random subset for $t$ and $z$, so that $\left\{\Delta \boldsymbol{h}^{t,z}\right\}$ captures the impact from a diverse set of randomly selected longer subsequences that span the entire input time 
series. We refer to this as the memory-discrepancy distillation loss for longer subsequences, denoted as $L_{\mathrm{MemKD-Long}}$.
We examined the effects of these two terms separately and found that combining them yields the best performance (see \cref{ablation}).

\subsubsection{Seamless Feature Distillation. }\label{feature_distillation}
While hidden states in RNNs capture temporal dependencies and information flow across time steps, they are internal representations of the data that serve a similar purpose to intermediate feature maps in temporal direction. Feature distillation refers to providing hints from intermediate representations to improve the training of the student, besides directly matching output distributions. Initially introduced in Fitnets~\cite{romero_fitnets_2014}, feature distillation loss is typically formulated as:
\begin{equation}\label{eq_fkd}
L_{\mathrm{Fea\_KD}} = \ell_\mathrm{f}(\phi_t(\psi _t(\boldsymbol{X})),\phi_s(\psi _s(\boldsymbol{X})))
\end{equation}
where $\psi _t(\boldsymbol{X})$ and $\psi _t(\boldsymbol{X})$ are feature maps of teacher and student respectively. Moreover, $\phi_t$, $\phi_s$ are some transformation functions 
to match the shape of teacher and student feature maps and $\ell_\mathrm{f}(.)$ is a similarity measure to match feature maps.
When we consider the hidden state matrices $\boldsymbol{H}_t \in \mathbb{R}^{(T \times m_t)}$ and $\boldsymbol{H}_s \in \mathbb{R}^{(T \times m_s)}$ across all temporal steps as $\psi_t(\boldsymbol{X})$ and $\psi_s(\boldsymbol{X})$, respectively, and $f_t$ and $f_s$ (in \Cref{eq_ft} and \Cref{eq_fs}) as $\phi_t$ and $\phi_s$, respectively, our proposed method can be seen as a feature distillation method.

Generally it is common for the student's intermediate feature dimensionality to be smaller than that of the teacher's. To bridge this gap, many methods~\cite{romero_fitnets_2014,huang_like_2017} introduce an intermediate fully connected layer to connect hint (feature layer of teacher) and guided (feature layer of student) layers. However, this additional layer introduces numerous auxiliary parameters, which significantly increase the space and computational demands while training, sometimes even surpassing the size of both teacher and student models~\cite{chen_learning_2021}.
In contrast, our proposed method does not necessitate such a regressor, even when the shapes of our feature maps differ. We calculate the feature distillation loss in an indirect yet semantic manner, thereby avoiding the need for explicit alignment of feature map dimensions. Therefore, our method stands out as an efficient feature distillation method that imposes no additional computational burden to align features.

\section{Experiments}\label{experiments}
\subsection{Teacher and Student Models}
In our experiments, we primarily focus on LSTM, a variant of RNN that incorporates a modified hidden state update mechanism, making it more effective in capturing long-term dependencies. These models are extensively employed in practical applications. We selected a three-layer LSTM model, with each layer having a hidden size of 100, as our teacher model. 
For our primary experiments, we used a single-layer LSTM model with a hidden size of 8 (denoted SL1\_8) as our student model. Unless otherwise specified, we refer to this model as the student. To evaluate the performance of our method under various levels of compression, we constructed additional versions of student models by varying the number of layers and hidden sizes. The total number of parameters, model sizes, and compression ratios with respect to the teacher model for all constructed models are summarized in~\Cref{tab:teacher_student_config}. Our primary student model achieves a compression factor of approximately 550 compared to the teacher model.
\networkParams

\subsection{Datasets} 

We used 56 datasets from UCR-2015 archive~\cite{chen_ucr_2015}, which consists of an extensive catalog of time series from various domains. We selected datasets representing three categories based on time series length: 'short', 'medium', and 'long'~\cite{Xiao_2021_new, xing_2022_efficient, campos_2023_lightts}. The length of a 'short' dataset is no more than 150, a 'medium' dataset varies from 150 to 500, and a 'long' dataset has a length of over 500.
Additionally, following~\cite{campos_2023_lightts}, we prioritized datasets with a large number of output classes, as these are considered the most challenging and have significant real-world applications. Our final set includes 23 short, 17 medium, and 16 long datasets, of which 18 have five or more output classes. The datasets ensure a comprehensive evaluation of TSC methods across various domains, encompassing different recording durations, sensor configurations, and data collection methods. An overview of these datasets is provided in Table~\ref{tab:datasets}. To ensure a fair evaluation, we maintained the original train/test split and allocated 20\% of the training set as a validation set. As a preprocessing step, we standardized all time series to a length of 100 using linear interpolation, followed by z-normalization.\\

\datasetsNew

\subsection{Experiment Setup}
For each dataset, we first train a teacher model across five random initializations and select the best-performing model for distilling knowledge to student models. The best teacher model is chosen based on the area under the precision-recall curve (AUC-PRC) evaluated on the validation set, ensuring reproducibility in real-world settings~\cite{ay_study_2022}. Next, we train the student models by distilling knowledge from the selected teacher model. In addition to MemKD, we implement state-of-the-art distillation methods for comparison, including Base-KD~\cite{hinton_distilling_2014}, FitNet~\cite{romero_fitnets_2014}, RKD~\cite{park_relational_2019}, Att~\cite{zagoruyko_paying_2017}, DKD~\cite{zhao_decoupled_2022}, and DT2W~\cite{qiao_class-incremental_2023}. To serve as a baseline, we also train identical student models (\textit{Base}) from scratch without knowledge distillation.
For all student models involving KD, we trained them using a combination of the distillation loss and the classification loss (cross-entropy loss) as:
\begin{equation}
    L_{\mathrm{train}} = \alpha\times L_{\mathrm{CE}} + \beta\times L_{\mathrm{KD}}
    \label{total_loss}
\end{equation}
where $\alpha$ and $\beta$ decide the contribution of classification loss $L_{\mathrm{CE}}$ and distillation loss $L_{\mathrm{KD}}$ for the total train loss $L_{\mathrm{train}}$, respectively.
For all experiments, $\alpha$ is fixed at 1, while the optimal value of $\beta$ is selected via grid search over ${0.01, 0.1, 0.5, 1, 10, 100, 150, 200, 300}$.
Each student model is trained for five random initializations, and evaluation metrics are reported as the average across these five runs. All models are implemented in Python using the PyTorch framework~\cite{paszke_2019_pytorch} following the same setting. Training is performed with the Adam optimizer, a batch size of 32, and a maximum of 500 epochs, with early stopping based on validation-loss and a patience of 50 epochs. A learning rate decay factor of 0.5 is applied at epochs 25, 30, and 35 for all models. For SL1\_4, SL1\_8, and SL1\_16 models, the initial learning rate is set to 0.1, while for the teacher and other student variants, it is set to 0.01.

We evaluated the performance of each model using three widely adopted metrics: area under the receiver operating characteristic curve (AUC-ROC), average AUC-PRC, and accuracy on the test set. The decision threshold for calculating accuracy was set at 0.5. For comparison, we employed a win/tie/loss calculation, where a model is considered to 'win' on a particular dataset if it achieves a higher AUC-PRC than the other models. We prioritized AUC-PRC over AUC-ROC and accuracy because AUC-PRC is more robust to class imbalance, a common characteristic among UCR datasets. Additionally, we calculated the average rank across datasets for each model.
Since we used averaged values for the student models, comparisons with teacher models were also made using their averaged values over five runs (instead of top-1) where necessary.


\section{Results and Discussion}
\subsection{Comparison with baseline methods}
\BasevsOursCombined
We first evaluated how the proposed MemKD performs compared to baseline approaches. Specifically, we compared MemKD's performance with that of a student trained from scratch (Base) and vanilla knowledge distillation (Base-KD). 
The results of the comparison across all 56 datasets are summarized in~\cref{tab:base_vs_ours}. It is evident that our method outperforms the Base model on all 56 datasets, establishing our proposed method as an efficient knowledge distillation framework. Additionally, the average performance metrics improved by approximately 10\% compared to the Base model. This is further illustrated in~\cref{fig:base_ours}, which plots the AUC-PRC of MemKD against that of the Base model. MemKD also outperforms vanilla knowledge distillation on approximately 90\% of the datasets, as further depicted in~\cref{fig:hkd_ours}.
Since MemKD is also a feature distillation method (see \cref{feature_distillation}), we compared it with vanilla feature distillation (Fitnets), and the summarized results are presented in~\cref{tab:fits_vs_ours}. MemKD outperforms Fitnets on 51 out of 56 datasets, achieving a 5\% improvement in the average performance metrics compared to Fitnets. This is further illustrated in~\cref{fig:fits_ours}, where MemKD demonstrates superior performance in over 90\% of the cases. These results demonstrate that our loss function better exploits the knowledge from the teacher’s feature layers, rather than providing direct hints from feature layers, while also eliminating the need for additional regression layers.

\AUCPlots

\subsection{Comparison with state-of-the-art methods}
Table~\ref{tab:sota} presents the evaluation results of different KD variants across the 56 datasets. We included win/tie/loss calculations both with and without the teacher model. The proposed method, MemKD, consistently outperforms all other distillation objectives on nearly 50\% of the datasets. The average rank of MemKD is 2.29, which is significantly lower than that of other methods, indicating that MemKD delivers competitive performance across the remaining datasets as well. None of the other methods were able to win more than 20\% of the datasets, nor did they achieve ranks close to 2. Additionally, our method shows at least a 2\% improvement over other methods in terms of average AUC-PRC. All KD methods report lower ranks compared to the Base model, indicating that they all benefit from knowledge distillation. Furthermore, MemKD achieves the lowest average rank even in the calculations that include the teacher model. 
It is the only method to exceed the teacher in terms of average AUC-PRC and number of wins, as none of the other methods were able to do so.
\sotaFiftySixNew

\subsection{Comparison with Teacher model}
We analyze the performance of MemKD across various compression levels compared to the teacher model. The primary goal of knowledge distillation is to achieve substantial compression while maintaining performance close to that of the teacher. There is always a trade-off between performance and compression, and the appropriate balance depends on the application. \Cref{tab:diff_architect} summarizes the results, comparing the teacher model to 9 different MemKD-student variants, with compression factors ranging from 2.5x up to 1600x. For each variant, we also include results from the Base model, which has the same compression but is trained without knowledge distillation. The findings show that MemKD delivers performance comparable to the teacher across all 9 student variants. For the smallest student model, SL1\_4, wins between MemKD and the teacher are split roughly evenly. However, starting from SL\_8, MemKD surpasses the teacher in over 60\% of the datasets, and this percentage increases to nearly 90\% towards the larger student models. MemKD also exceeds the teacher's average AUC-PRC in all student variants except for the smallest one. In contrast, none of the Base model variants outperform the teacher, consistently losing in over 90\% of the datasets. This demonstrates that MemKD enables significant performance improvements across a range of compression levels that the scratch model cannot achieve.
\diffarchitect

\subsection{Ablation study}\label{ablation}

We conducted an ablation study using 28 datasets to evaluate the effectiveness of MemKD-Short and MemKD-Long, which are induced by memory retention over shorter and longer subsequences, respectively. To generate MemKD-Long, we selected 100 different pairs of values for $z$ and $t$, producing 100 distinct subsequences spanning the input time series. The ablation results, which highlight the importance of each component, are summarized in~\cref{tab:ablation}. We observed that MemKD-Long showed a slight improvement over MemKD-Short, with an increase of 1-2\% in average performance metrics. Moreover, the combination of both components outperforms the individual components, improving performance on over 50\% (15 out of 28) of the datasets, and achieving an average rank of 1.54 across the individual components and the baseline without any components.
We also tested the impact of varying the number of subsequences and found no significant benefit in increasing it beyond 100 (see \cref{tab:num_pairs}).
\ablation
\numPairs

\sensitivity
\subsection{Sensitivity Analysis on $\beta$. }

We also conducted an ablation study on $\beta$, which controls the importance of the memory-discrepancy distillation loss term relative to the task loss for six different datasets. The results are shown in~\Cref{pic:sensitivity}. AUC-PRC scores improved for almost all non-zero values of $\beta$ compared to when $\beta=0$ (i.e. when our loss term is totally absent). The performance of our model remains stable across moderate values of $\beta$ and is significantly better than when $\beta=0$. For very large values of $\beta$, the performance improvement over $\beta=0$ diminishes.
This is expected because when $\beta$ is too large, the distillation loss dominates the total loss, which can lead to over-regularization. 
In other words, the student model focuses excessively on mimicking the teacher model, potentially at the expense of learning the task-specific features necessary for optimal performance on the target task.

\TSNEPlots
\subsection{Evidence for Hidden-State Alignment with the Teacher}
\Cref{fig:tsne_plots} visualizes the last five hidden-state representations of the teacher, Base, Base-KD, and MemKD models on UWaveGestureLibraryAll using t-SNE. The Base student exhibits scattered and substantially overlapping hidden-state representations, indicating less discriminative placement of samples in the latent space. Conventional KD (Base-KD) improves this organization to some extent, but considerable inter-class overlap remains. In contrast, MemKD produces more compact and distinctly placed class-specific regions, more closely resembling the well-organized structure observed in the teacher. Since the visualization incorporates the last five hidden states rather than only the final representation, this result suggests that MemKD improves the placement and organization of representations across the student's overall hidden states, progressively forming more class-discriminative feature structures. This supports the intended role of MemKD in transferring teacher knowledge throughout the student's hidden-state learning process rather than constraining supervision only to the final representation.

\section{Conclusion}\label{conculsion}
Most prior work on feature distillation has explicitly focused on computer vision tasks, with loss functions assumed to capture knowledge solely about visual features. These methods often lack clear analogies for time series. We identified and addressed this limitation by introducing a novel KD framework tailored for deep time series recurrent networks. Specifically, we investigated how memory differs and evolves within RNNs and whether we can use information about memory retention over short contiguous subsequences to improve the performance of a student model. By leveraging a unique loss function that captures memory retention discrepancies across subsequences, MemKD effectively transfers knowledge from a large, complex LSTM model to a significantly smaller LSTM model. The extensive experiments, using fifty-six time series datasets, demonstrate that MemKD significantly outperforms logit-level and feature-level state-of-the-art KD methods, achieving substantial reductions in parameter size and memory usage while maintaining comparable performance. Future work will focus on further optimizing the MemKD framework and studying the knowledge of memory retention on longer subsequences, providing comprehensive coverage over the whole time series instance.

\section*{Acknowledgments}
This research was supported by The University of Melbourne’s Research Computing Services and the Petascale Campus Initiative. 
We also gratefully acknowledge the authors and maintainers of the UCR time series repositories for making their datasets publicly available.

\section*{Declaration of generative AI and AI-assisted technologies in the manuscript preparation process}
In preparing this manuscript, we used OpenAI’s ChatGPT (GPT-4) as a writing assistance tool. 
Its role was limited to polishing language for improved clarity, grammar, and readability in certain sections of the paper. 
The model did not contribute to research ideation, methodological design, experimental execution, data analysis, or interpretation of results. 
All scientific content, technical contributions, and conclusions are solely the work of the authors. 
We take full responsibility for the accuracy and integrity of the paper.

\section*{Data Availability Statement}
The datasets used in this study are publicly available from the repositories cited and linked in the manuscript. No new datasets were generated as part of this study.





\bibliographystyle{elsarticle-num-names}
\bibliography{ref}






\end{document}